\documentclass[letterpaper, 10 pt, conference]{ieeeconf}  

\IEEEoverridecommandlockouts                              

\usepackage{graphics} 
\usepackage{epsfig} 
\usepackage{amsmath} 
\usepackage{listings}

\usepackage{cite} 
\usepackage{units}
\usepackage{color}
\usepackage{multirow}

\newcommand{\hl}[1]{\textcolor{blue}{#1}}

\renewcommand{\hl}[1]{#1}

\title{\LARGE \bf Reconsidering Fascicles in Soft Pneumatic Actuator Packs}
\author{Wyatt Felt
\thanks{Email: {\tt\small wfelt@umich.edu}}%
}

\begin{document}

\maketitle
\thispagestyle{empty}
\pagestyle{empty}

\begin{abstract}
This paper discusses and contests the claims of ``Soft Pneumatic Actuator Fascicles for High Force and Reliability'' a research article which was originally published in the March 2017 issue of the Journal \textit{Soft Robotics}. The original paper claims that the summed forces of multiple thin-walled extending McKibben muscles are greater than a volumetrically equivalent actuator of the same length at the same pressure. The original paper also claims that the purported benefit becomes more pronounced as the number of smaller actuators is increased. Using reasonable assumptions, the analysis of this paper shows that the claims of the original paper violate the law of conservation of energy. This paper also identifies errors in the original methodology that may have led to the erroneous conclusions of the original paper. The goal of this paper is to correct the record and to provide a more accurate framework for considering fascicles used in soft pneumatic actuator packs.
\end{abstract}

\section{Introduction}
The March 2017 issue of the journal \textit{Soft Robotics} included a research article \cite{fasciclesOriginal} that considered the use of multiple Soft Pneumatic Actuators (SPAs) in parallel combinations. While the original paper is interesting in many ways, the central claims of the paper appear to be based on faulty analysis. The present paper seeks to provide a more thorough analysis, to correct some of the main conclusions of the original, and, thereby, to provide a sound theoretical basis for the design of pneumatic actuator packs with multiple individual actuators working in parallel.

\begin{figure}
    \centering
    \includegraphics[width = 3.1 in]{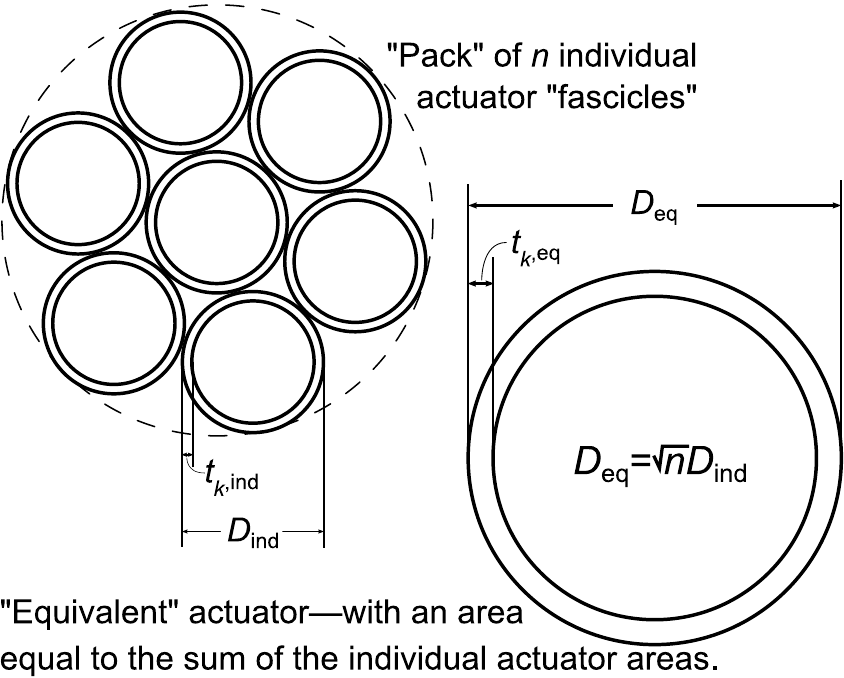}
    \caption{The original paper makes the erroneous claim that the force output from a pack of $n$ individual actuators exceeds that of a single volumetrically comparable actuator. The paper further claims that this benefit becomes more pronounced as $n$ increases.}
    \label{fig:IndEq}
\end{figure}

The original paper \cite{fasciclesOriginal} has several interesting contributions. It explores combining extending McKibben muscle actuators together to sum their forces. Each individual actuator is referred to as a ``fascicle'' and together they form a ``pack'' (Fig.~\ref{fig:IndEq}). The pack configuration is useful to achieve high forces without increasing the size of the individual actuators. By using four actuators, for example, the force output of the pack is quadruple that of the individual fascicles. The authors also show how the relatively small individual actuators can be arranged in patterns that make them less bulky. For instance, arranging the actuators in a row creates a low-profile pack that could be worn under clothing. This flat pack is certainly less cumbersome than a single actuator with a larger diameter.

\hl{Though} \cite{fasciclesOriginal} \hl{considers \textit{extending} McKibben muscles, others have looked at effect of using multiple, thin \textit{contracting} McKibben muscles. Suzumori's group, for example, has been experimenting with bundles of thin contracting McKibben muscles since at least 2011} \cite{Wakimoto2011,Kuramaya2017}. \hl{Their work shows that, due to the large diameter change in contracting McKibben muscles, a ``mulifilament'' muscle can exhibit a greater contraction ratio than the individual actuators} \cite{Kuramaya2017}. \hl{Others have looked at how the principles of ``variable recruitment'' (which is common in mammalian muscle) can be applied to groups of contracting McKibben muscles to achieve gains in efficiency} \cite{delahunt2016,meller2016improving,bryant2014variable,robinson2015variable,chapman2018bioinspired}. 

Though not without merit, the original paper,\cite{fasciclesOriginal}, presents specious claims about the benefits of fascicles. The central claim could be summarized as follows: \textit{By using multiple thin-walled soft pneumatic actuator ``fascicles,'' it is possible to achieve higher forces in a pack than a single larger actuator with an equivalent length and volume at the same pressure could produce}. The original paper further claims that the purported improvement in volumetric force-density increases with increasing numbers of fascicles. These counter-intuitive (and thus attractive) claims appear to be based on a faulty analysis of the McKibben muscle force equations. 

The analysis presented in this work demonstrates that the claims in the original paper are not true.
\hl{The primary contribution of this paper is} to provide a more thorough and correct analysis of the models incorrectly used in \cite{fasciclesOriginal}. It appears that errors in the analysis of these models led to the erroneous claims of the original paper. 

\section{Analysis}

\subsection{Original claim}

The original paper claims that the sum of forces from the individual actuator ``fascicles'' in a pack is greater than the force from ``an equivalent single SPA of comparable volume.'' The original paper further predicts that this benefit would become more pronounced as the number of fascicles in the pack is increased. Examples of the claims in the paper include:

\hl{From the abstract,}
\begin{quote} ``Experimental measurements show an SPA pack to generate over \unit[112]{N} linear force, while the model indicates the benefit of parallel actuator grouping over a geometrically equivalent single SPA scale as an increasing function of the number of individual actuators in the group. For a module of four actuators, a \unit[23]{\%} increase in force production over a volumetrically equivalent single SPA is predicted and validated, while further gains appear possible up to \unit[50]{\%}.'' \end{quote} 

From the methods, \begin{quotation} ``By grouping SPAs in parallel, actuator packs can be formed, which outperform individual SPAs of comparable size. Not only can higher force output be achieved with this configuration but also the benefit mutually increases with the number of constitutive actuator units utilized in the pack. This is to say that the more unit actuators are combined in parallel, the greater the gain in force output can be attained. This improvement in spatial-force efficiency allows for either stronger actuation in a given physical space or more compact actuation for a given required force.''\end{quotation}

And from the discussion (emphasis original), \begin{quotation} ``It is shown in the Results section that \textit{a fascicle arrangement of SPAs is capable of generating more linear force than an equivalent single SPA of comparable volume.} Moreover, the gain in force production continues to increase within practical bounds as the number of units in a parallel configured SPA pack is increased. This model-based finding indicates that high-performance SPA design favors multiplicity and can be exploited as a new actuator design strategy, which takes into consideration the effect of multiple unit actuators in parallel coordination as well as the individual unit actuator design parameters to achieve desired performance.'' \end{quotation}

\hl{With regard to the word ``validated'' in the abstract,} \cite{fasciclesOriginal} \hl{reports no experimental comparison between a ``volumetrically equivalent single SPA'' and a ``module of [n] actuators.'' The force output of a four-actuator pack \textit{is} tested experimentally, but the force data from this pack are not compared against forces from a volumetrically equivalent actuator. The predicted ``}\unit[23]{\%} \hl{increase in force [from a four actuator pack compared to a single equivalent actuator]'' is ``validated'' only against an erroneous prediction from the model.}

The erroneous claims of \cite{fasciclesOriginal} have been repeated in subsequent publications. For example, \cite{Tawk2018} states
\begin{quotation} ``[The authors of \cite{fasciclesOriginal}] have proved that the arrangement of multiple smaller soft pneumatic actuators produces larger output forces compared to one soft actuator with the same total volume.''
\end{quotation}

Because the original paper uses only a simple model to establish its claims, this paper uses only the same simple model (correctly analyzed) to refute it. This paper presents no challenge to the experimental data of the original paper, only to the paper's untested predictions and conclusions.

\subsection{Actuator model}

This work seeks to mirror the assumptions of the original paper insofar as possible without duplicating errors. The model used in \cite{fasciclesOriginal} was presented by Chou and Hannaford \cite{ChouHannaford} and is common in the literature. The model assumes that the McKibben muscle is cylindrical and that its kinematics are governed by an inextensible fiber braid. The outer dimensions of the actuator are given by the axial length $L$ and diameter $D$ of the braid
\begin{equation}
L = b \cos{\theta}
\label{eq:length}
\end{equation}
\begin{equation}
D = \frac{b \sin{\theta}}{N \pi} 
\label{eq:diameter}
\end{equation}
where $b$ is the unwound length of the fibers in the braid, $N$ is the number of times they circle the axis and $\theta$ is a measure of their angle with respect to the long axis.

The diameter $D$ can also be expressed as a fraction of $D_0$, where $D_0 = \frac{b}{N\pi}$ is the theoretical diameter of the fibers if the braid angle were \unit[90]{$^\circ$} (i.e. $\theta = \pi/2$)
\begin{equation}
D = D_0 \sin{\theta} .
\label{eq:D_from_D0}
\end{equation}

The values of $D_0$, $N$ and $b$ are determined by the fabrication of the actuator. Unlike the length $L$, diameter $D$ and braid angle $\theta$, they do not change as the actuator creates motion. 

Neglecting the wall-thickness of the actuator (thin-walled approximation), the force output $F_{\text{thin}}$ of the actuator is approximated in \cite{ChouHannaford} as
\begin{equation}
    F_{\text{thin}} = \frac{\pi P'}{4} {D_0}^2 \left( 3 \cos^2\!\left(\theta\right) - 1 \right)
    \label{eq:F_thin}
\end{equation}
where P' is the internal gauge pressure of the actuator.

This equation is presented in \cite{ChouHannaford} for \textit{contractile} McKibben muscles (i.e. $\theta < 54.7^\circ$). Accordingly, a positive value of $F_{\text{thin}}$ corresponds to a tensile force. The \textit{extending} McKibben muscles considered in this work and in \cite{fasciclesOriginal} thus have braid angles larger than \unit[54.7]{$^\circ$} and correspondingly negative values of $F$. Nevertheless, when a force is said to be ``higher'' or have an ``increase'' compared to another, it is understood that it is larger in magnitude.

The analysis in \cite{fasciclesOriginal} is based on a form of the equation in \cite{ChouHannaford} that takes into account the thickness $t_k$ of the elastomeric actuator walls. For clarity, the force output predicted from this approximation is designated $F_{\text{thick}}$ here. In \cite{ChouHannaford} it has the following form
\begin{equation}
\begin{aligned}
      F_{\text{thick}} = &\frac{\pi P'}{4} {D_0}^2 \left( 3 \cos^2\!\left(\theta\right) - 1 \right) \\ 
      &+ \pi P' \left( D_0 t_k \left(2\sin{\theta} - \frac{1}{\sin{\theta}} \right) - {t_k}^2 \right) .
      \end{aligned}
      \label{eq:F_thick}
\end{equation}
This approximation takes into account the volumetric effect of the wall-thickness but not its elastic forces that resist the actuator motion.

When presented in \cite{fasciclesOriginal}, the authors use the identity of $D_0$ to rewrite the equation as follows
\begin{equation}
\begin{aligned}
      F_{\text{thick}} = &\frac{b^2 P'}{4 \pi N^2} \left( 3 \cos^2\!\left(\theta\right) - 1 \right) \\ 
      &+ \pi P' \left( \frac{b}{N \pi} t_k \left(2\sin{\theta} - \frac{1}{\sin{\theta}} \right) - {t_k}^2 \right) .
      \end{aligned}
\end{equation}
They also provide a valid expression for the fiber length $b$ based on the Pythagorean theorem and the circumference of the cylinder
\begin{equation}
    b = \sqrt{L^2 + \left(D\pi N\right)^2}.
    \label{eq:b_pyth}
\end{equation}


\subsection{Individual actuator parameters}

The individual actuator parameters presented in the original paper are not self-consistent. That is, they do not satisfy the model equations. This inconsistency is corrected in this work by identifying the value of $\theta$ that preserves the consistency of the rest of the parameters.

\begin{table}[htp]
    \centering
    \caption{Individual actuator parameter values as defined in the original paper (not self-consistent)}
    \begin{tabular}{cc}
\hline 
\textit{Parameter}  & \textit{Value} \\ \hline
$L$  & \unit[145]{mm} \\
$D$  & \unit[17]{mm} \\
$\theta$ & \unit[75.2]{$^\circ$} \\ 
$N$ & 16 \\ \hline
    \end{tabular}
    \label{tab:table1}
\end{table}
The parameters of the individual actuators given in the original paper are listed in Table \ref{tab:table1}. 
These values are not self-consistent. That is, using the listed parameters to calculate $b$ results in different values from each of the three equations, Eq.~\eqref{eq:length}, Eq.~\eqref{eq:diameter} and Eq.~\eqref{eq:b_pyth}.

Assuming that the easy-to-measure values of $L$, $D$ and $N$ are correct, this work uses Eq.~\eqref{eq:b_pyth} to calculate the value of $b$ as \unit[866.7]{mm}. This value of $b$ results in the same value of $\theta$ from both Eq.~\eqref{eq:length} and Eq.~\eqref{eq:diameter}, $\theta =$ \unit[80.4]{$^\circ$}. The resulting set of self-consistent parameters are listed in Table \ref{tab:table2}.
%
\begin{table}[htp]
    \centering
    \caption{Individual actuator parameter values considered here (self-consistent)}
    \begin{tabular}{cc}
\hline 
\textit{Parameter}  & \textit{Value} \\ \hline
$L$  & \unit[145]{mm} \\
$D$  & \unit[17]{mm} \\
$\theta$ & \textbf{\unit[80.369]{$^\circ$}} \\ 
$N$ & 16 \\ \hline
$b$ & \unit[866.728]{mm} \\
\hline
    \end{tabular}
    \label{tab:table2}
\end{table}

The original paper's parameters do not result in a physically meaningful braid description. This work uses the angle in Table~\ref{tab:table2} to avoid duplicating this small error in the original work.

\subsection{Generating ``equivalent'' actuator parameters}

A more serious problem in the original paper is found in the prescribed methods for generating the parameters of the ``equivalent'' actuator. Following the methodology described in the original paper results in another inconsistent parameter set for the equivalent actuator. By failing to adjust the the number of fiber turns with the change in diameter, the original paper appears to consider a geometrically impossible equivalent actuator.

The original paper claims that a pack with many fascicles can produce more force at the same pressure than an ``equivalent'' actuator of ``comparable volume.'' The volume is preserved by using an ``equivalent'' actuator with the same length as the pack but with a cross-sectional area that has been scaled to match the total cross-sectional area of the actuators in the pack.

The original paper states, \begin{quote} ``For a cross-sectional area, $A$, of an individual unit actuator, the ``equivalent SPA'' is defined to have equal volume to a pack of $n$ unit actuators, by defining its cross-sectional area as $A_\text{eq} = n A$. This simple relationship for the equivalent area defines the necessary diameter for an equivalent single SPA. All other actuator parameters, as shown in Table 1, are held equal.''\end{quote} 

Note that, in this work, the parameters of the ``individual unit actuator'' and ``equivalent'' actuators are designated with the subscripts ``ind'' and ``eq'' respectively.

The problem with the methodology described in the original paper is the holding of ``all other actuator parameters'' equal while $D$ is scaled. That is, problems arise when $L_{\text{ind}} = L_{\text{eq}}$, $\theta_{\text{ind}} = \theta_{\text{eq}}$, $N_{\text{ind}} = N_{\text{eq}}$ and $D_{\text{ind}} \neq D_{\text{eq}}$. Attempting to generate a set of equivalent braid parameters in this way results in a set that is not self-consistent. 

The number of turns $N$ must be defined differently if the diameter of the actuator is different and the other parameters are kept constant. It is possible that this error in calculation led the authors of the original paper to their erroneous conclusions.

It is reasonable to hold the length and braid angles constant (i.e. $L_{\text{ind}} = L_{\text{eq}}$ and $\theta_{\text{ind}} = \theta_{\text{eq}}$). Indeed, for the volume to be ``comparable'' with equivalent cross-sectional areas, the lengths must be equal. Similarly, the braid angle, which changes over the actuation stroke, largely dictates the force output of a McKibben muscle actuator \cite{ChouHannaford} (see also Eq.~\eqref{eq:F_thin} and Eq.~\eqref{eq:F_thick} in this work). Two McKibben muscle actuators with the same outer dimensions but different braid angles would produce different forces for the same pressure. To compare apples to apples, as they say, the braid angle must be the same in the equivalent actuator.

With these two parameters equal, Eq.~\eqref{eq:length} makes it clear that the length $b$ of the fibers in the individual and equivalent actuators must also be equal
\begin{equation}
    \begin{aligned}
 \text{Eq.~\eqref{eq:length}:  }   l &= b \cos{\theta} \\
    \text{Assuming } l_{\text{ind}} &= l_{\text{eq}} \\
    b_{\text{ind}} \cos{\theta_{\text{ind}}} &= b_{\text{eq}} \cos{\theta_{\text{eq}}} \\
    b_{\text{ind}} \cos{\theta_{\text{ind}}} &= b_{\text{eq}} \cos{\theta_{\text{ind}}} \\
    b_{\text{ind}} &= b_{\text{eq}}.
    \end{aligned}
    \label{eq:b_ind_equal_eq}
\end{equation}
If $b_{\text{ind}} = b_{\text{eq}}$ and $\theta_{\text{ind}} = \theta_{\text{eq}}$, then the number of fiber wraps in the two actuators are equal (i.e. $N_{\text{ind}} = N_{\text{eq}}$) \textit{if and only if} the diameters are equal (i.e. $D_{\text{ind}} = D_{\text{eq}}$)
\begin{equation}
    \begin{aligned}
 \text{Eq.~\eqref{eq:diameter}:  }   D &= \frac{b \sin{\theta}}{N \pi} \\
    \text{Assuming } D_{\text{ind}} &\neq D_{\text{eq}} \\
    \frac{b_{\text{ind}} \sin{\theta_{\text{ind}}}}{N_{\text{ind}} \pi} &\neq \frac{b_{\text{eq}} \sin{\theta_{\text{eq}}}}{N_{\text{eq}} \pi} \\
  \frac{b_{\text{ind}}\sin{\theta_{\text{ind}}}}{N_{\text{ind}} \pi} &\neq \frac{b_{\text{ind}}\sin{\theta_{\text{ind}}}}{N_{\text{eq}} \pi} \\
    N_{\text{ind}} &\neq N_{\text{eq}} \\.
    \end{aligned}
    \label{eq:N_iff_D}
\end{equation}
Thus, if $L$ and $\theta$ are kept the same, it is unreasonable to hold $N$ constant while scaling the diameter $D$ (as the original paper does). 

\begin{figure}
    \centering
    \includegraphics[width = 3.4 in]{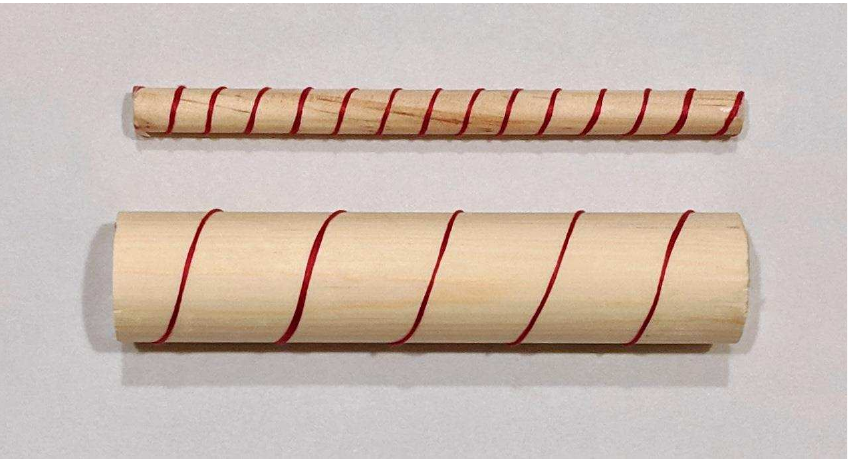}
    \caption{The original paper evaluates a larger diameter actuator while holding the length, braid angle and the number of turns of the fiber equal. This is not possible in a physical actuator. The number of fiber turns must change with the diameter. This can be demonstrated with a simple household experiment. Shown are two equal-length strings wrapped around the same axial-length of cylinders with different diameters. The angles of the fibers with respect to the cylinder axis are approximately equal. It is clear, however, that the fiber wraps around the thin cylinder more times than the larger-diameter cylinder.}
    \label{fig:wraps}
\end{figure}

This relationship between $D$ and $N$ can be easily verified by wrapping and taping two equal-length strings around two respective household cylinders of different diameters (Fig.~\ref{fig:wraps}). If the strings are wrapped around the same axial length of the cylinders, their angle with respect to the axis will be approximately equal. If the diameters of the cylinders are different, the number of turns around the cylinder will be different.

\subsection{Considering the claims without wall-thickness}

In this work, the original paper's problematic claims are first discussed without taking the elastomeric wall-thickness into account. The neglect of the wall-thickness is a reasonable approximation for actuators whose walls are thin relative to their diameters. To indicate that the wall-thickness has been neglected, the variables are marked with the subscript ``thin.''

Neglecting the wall-thickness, the external cross-sectional area $A_{\text{eq}}$ of the equivalent actuator is $n$-times greater than that of the individual actuator $A_{\text{ind}}$. The outer diameter $D_{\text{eq}}$ of the equivalent actuator is the product of the square-root of the number of actuator fascicles $n$ and the outer diameter of the individual actuator $D_{\text{ind}}$
\begin{equation}
\begin{aligned}
    A_{\text{eq}} &= n A_{\text{ind}} \\
    \left(\pi/4\right){D_{\text{eq}}}^2 &= n \left(\pi/4\right){D_{\text{ind}}}^2 \\
    D_{\text{eq}} &= \sqrt{n} D_{\text{ind}}.
    \end{aligned}
    \label{eq:D_scaling_A_ext_equal}
\end{equation}

The maximum diameter $D_0$ of the equivalent actuator scales in the same way as $D$. This can be seen by examining Eq.~\eqref{eq:D_from_D0} while considering that the braid angles between the two actuators are the same
\begin{equation}
 D_{0\text{,eq}} =\frac{D_{\text{eq}}}{D_{\text{ind}}} D_{0\text{,ind}} .
 \label{eq:D_0_scaling}
\end{equation}
 
By Eq.~\eqref{eq:D_scaling_A_ext_equal} and Eq.~\eqref{eq:D_0_scaling}, the maximum diameter of the equivalent actuator $D_{0\text{,eq}}$ is similarly scaled from that of the individual actuator $D_{0\text{,ind}}$
\begin{equation}
    D_{0\text{,eq}} = \sqrt{n} D_{0\text{,ind}} .
    \label{eq:D_0_thin_eq}
\end{equation}

\subsubsection{Method 1, Force Equation}

The total force output $F_{\text{thin,pack}}$ from a ``pack'' of identical individual fascicles in parallel is the sum of their individual forces given by Eq.~\eqref{eq:F_thin} (recall that the lowercase ``$n$'' is the count of the fascicles in the pack)
\begin{equation}
\begin{aligned}
    F_{\text{thin,pack}} &= n F_{\text{thin,ind}} \\
    F_{\text{thin,pack}} &= n \frac{\pi {\left( D_{0\text{,ind}} \right)}^2 P'}{4} \left( 3 \cos^2\!\left(\theta_{\text{ind}}\right) - 1 \right). 
    \end{aligned}
    \label{eq:F_pack_thin}
\end{equation}

The force output from the single equivalent actuator is also given by Eq.~\eqref{eq:F_thin}
\begin{equation}
    F_{\text{thin,eq}} = \frac{\pi {\left( D_{0\text{,eq}} \right)}^2 P'}{4} \left( 3 \cos^2\!\left(\theta_{\text{eq}}\right) - 1 \right).
\end{equation}
Substituting for $D_{0\text{,eq}}$ from Eq.~\eqref{eq:D_0_thin_eq} and considering that $\theta_{\text{eq}} = \theta_{\text{ind}}$, the force from the equivalent actuator is
\begin{equation}
\begin{aligned}
     F_{\text{thin,eq}} &= \frac{\pi {\left( \sqrt{n} D_{0\text{,ind}} \right)}^2 P'}{4} \left( 3 \cos^2\!\left(\theta_{\text{ind}}\right) - 1 \right) \\
     F_{\text{thin,eq}} &= n \frac{\pi {\left( D_{0\text{,ind}} \right)}^2 P'}{4} \left( 3 \cos^2\!\left(\theta_{\text{ind}}\right) - 1 \right).
    \end{aligned}
    \label{eq:F_eq_thin}
\end{equation}
Eq.~\eqref{eq:F_eq_thin} is equivalent to Eq.~\eqref{eq:F_pack_thin}.

Thus, when neglecting the wall-thickness, the force from an equivalent actuator at the same pressure is predicted to be identical to the force from the actuator pack, regardless of the number of fascicles used
\begin{equation}
    F_{\text{thin,eq}} = F_{\text{thin,pack}}.
\end{equation}
This contradicts the claims of the original paper.

\subsubsection{Method 2, Conservation of Energy}

The law of conservation of energy can also be used to evaluate the claims. 

Consider an ideal actuator without losses or energy storage. The input energy $E_{\text{in}}$ to the actuator is equal to the output energy $E_{\text{out}}$
\begin{equation}
\begin{aligned}
     E_{\text{in}} &= E_{\text{out}} \\
     P'{\Delta}V &= F_{\text{avg}}{\Delta}L.
     \label{eq:EinEout}
\end{aligned}
\end{equation}
The input is flow energy, that is, the product of the gauge pressure in the actuator $P'$ (considered to be constant) and the change in volume ${\Delta}V$. The output energy is mechanical work, that is, the product of the average extension force $F_{\text{avg}}$ and the distance over which that force is applied (written as a change of length ${\Delta}L$) \cite{FeltRoboSoft2018}.

Considering two closed-volume fluid-powered actuators actuated by the same constant pressure $P'$ and acting over the same distance ${\Delta}L$, the average force of one actuator can only exceed the other if it undergoes a larger change in volume ${\Delta}V$. For the claims of the original paper to be valid, the sum of the volume changes of the individual actuators would need to be larger than the volume change of the equivalent actuator. This, however, is not the case.

To illustrate this, consider an extending McKibben muscle at an initial state before \hl{extension} (subscript ``1'') and final state (``2'') after \hl{extension}. The change of volume ${\Delta}V$, with the same assumptions as the force model, is 
\begin{equation}
\begin{aligned}
{\Delta}V &=  V_2 - V_1 \\
 &= \frac{\pi}{4}\left({D_2}^2{L_2} - {D_1}^2{L_1} \right) .
\end{aligned}
\label{eq:Delta_V}
\end{equation}

\hl{Though the actuator extends axially, the value of} $D_0$ \hl{remains the same.}
\hl{Accordingly,} Eq.~\eqref{eq:D_from_D0} \hl{can be re-arranged to find the relationship between the relaxed diameter} $D_1$ \hl{and the post-extension diameter} $D_2$. $\theta_1$ and $\theta_2$ are the respective fiber angles in the relaxed and extended states.
\begin{equation}
\begin{aligned}
D_0 = \frac{D_1}{\sin{\theta_1}} &= \frac{D_2}{\sin{\theta_2}} \\
{D_2} &= \frac{\sin{\theta_2}}{\sin{\theta_1}}{D_1} \\
{D_2} &= \gamma{D_1} \\
\gamma &= \frac{\sin{\theta_2}}{\sin{\theta_1}}.
\end{aligned}
\label{eq:gamma}
\end{equation}
\hl{When the actuator extends from} $L_1$ \hl{to} $L_2$, $D_2$ is scaled down from the initial diameter $D_1$ by a factor of $\gamma$.

\hl{The value of} $\gamma$ \hl{depends on} $\theta_1$ and $\theta_2$ \hl{which, in turn, can be found with length of the actuator} $L$ \hl{at the respective angles and the constant fiber-length} $b$ (Eq.~\eqref{eq:length}). \hl{Because the individual and equivalent actuators in this analysis have the same values for} $L_1$, $L_2$ and $b$, \hl{they share the same values of} $\theta_1$, $\theta_2$ and $\gamma$.

With Eq.~\eqref{eq:gamma}, the change in volume from Eq.~\eqref{eq:Delta_V} is
\begin{equation}
\begin{aligned}
{\Delta}V &= \frac{\pi}{4}\left({D_2}^2{L_2} - {D_1}^2{L_1} \right) \\
&= \frac{\pi}{4}\left({\left( \gamma D_1 \right)}^2{L_2} - {D_1}^2{L_1} \right) \\
&= \frac{\pi}{4}{D_1}^2\left({\gamma}^2{L_2} - {L_1} \right).
\end{aligned}
\label{eq:Delta_V_gamma}
\end{equation}
The total change in volume for the pack of $n$ actuators ${\Delta}V_{\text{pack}}$ is $n$-times the change of the individual actuators ${\Delta}V_{\text{ind}}$
\begin{equation}
\begin{aligned}
{\Delta}V_{\text{pack}} &= n {\Delta}V_{\text{ind}} \\
&= n \frac{\pi}{4}{D_{1\text{,ind}}}^2\left({\gamma}^2{L_2} - {L_1} \right).
\end{aligned}
\end{equation}
By Eq.~\eqref{eq:D_scaling_A_ext_equal}, the change in volume for the equivalent actuator ${\Delta}V_{\text{eq}}$ is
\begin{equation}
\begin{aligned}
{\Delta}V_{\text{eq}} &= \frac{\pi}{4}{D_{1\text{,eq}}}^2\left({\gamma}^2{L_2} - {L_1} \right) \\
&= \frac{\pi}{4}{\left( \sqrt{n} D_{1\text{,ind}} \right)}^2\left({\gamma}^2{L_2} - {L_1} \right) \\
&= n \frac{\pi}{4}{D_{1\text{,ind}}}^2\left({\gamma}^2{L_2} - {L_1} \right).
\end{aligned}
\end{equation}
The change volume is the same
\begin{equation}
 {\Delta}V_{\text{pack}} =  {\Delta}V_{\text{eq}}.   
\end{equation}
By Eq.~\eqref{eq:EinEout}, which assumes ideal actuators, the average force from a pack of individual actuators $F_{\text{avg,pack}}$ is predicted to be equal to the average force from an equivalent actuator $F_{\text{avg,eq}}$ at the same pressure, undergoing the same length change
\begin{equation}
 F_{\text{avg,pack}} = F_{\text{avg,eq}} .
\end{equation}
This is true regardless of how many actuators are used to form the pack. Accordingly, it contradicts the claims of the original paper.

\subsection{Assumptions about wall-thickness}

It has been shown, without considering the wall-thickness, that there is no force benefit from increasing the number of fascicles (compared to an equivalent actuator). Both the force equation used by the original work and the conservation of energy contradict the claims of the original work. This work now considers the wall-thickness and shows, under reasonable assumptions, \hl{that the addition of wall-thickness does add support for the claims of the original paper.}

One thing that is not explored in the original work is the effect of the elastomeric bladder inside the McKibben muscle. The walls of the bladder create non-trivial elastic forces that resist actuator motion and reduce the output forces. These elastic forces were not considered in the original paper and they are not considered here. What \textit{is} considered in the original paper is the effect of the wall-thickness $t_k$ on the internal volume of the actuator. Because the force output of the actuator is related to the rate of volume change, this thickness term $t_k$ also appears in the force equation Eq.~\eqref{eq:F_thick}. Its effect is to reduce the force-output of the actuator.

The original paper does not discuss the effect of the wall-thickness on the definition of the ``equivalent'' actuator. An increased wall-thickness reduces the internal cross-sectional area of the actuator. The original paper does not specify whether the cross-sectional area used to define equivalence is the internal or external cross-sectional area. However, the original paper \textit{does} state that the the equivalent actuator has a ``comparable volume'' to total volume of individual actuators in the pack. Accordingly, this work assumes that the \textit{external} cross-sectional area of the equivalent actuator is equal to the total \textit{external} cross-sectional area of the individual actuators in the pack. This assumption makes the diameter relationship between the equivalent and individual actuators identical to the relationship found in Eq.~\eqref{eq:D_scaling_A_ext_equal} (i.e. $D_{\text{eq}} = \sqrt{n}D_{\text{ind}}$).

The fraction of the external diameter $D$ made up by the thickness of each wall $t_k$ is designated $\hat{t}_k$
\begin{equation}
    \begin{aligned}
    0 \leq \hat{t}_k \leq 0.5 \\
    \hat{t}_k = \frac{t_k}{D} .
    \end{aligned}
\end{equation}
This quantity is bounded between zero (no wall thickness) and one half (no hollow interior). 

For the sake of simplicity, this paper assumes that the relative wall-thickness of the equivalent actuator $\hat{t}_{k\text{,eq}}$ is equal to the relative wall-thickness of the individual actuator $\hat{t}_{k\text{,ind}}$. That is, if the thickness of the original actuator walls made up \unit[10]{\%} of the overall diameter, the walls of the equivalent actuator would make up \unit[10]{\%} of the equivalent actuator's overall diameter
\begin{equation}
\begin{aligned}
    \hat{t}_{k\text{,ind}}  &= \hat{t}_{k\text{,eq}} \\
    \frac{t_{k,\text{ind}}}{D_{\text{ind}}} &= \frac{t_{k,\text{eq}}}{D_{\text{eq}}}.
\end{aligned}
\end{equation}
\hl{When} $\hat{t}_{k\text{,ind}} = \hat{t}_{k\text{,eq}}$, \hl{the equivalent actuator has the same total volume of elastomeric material as the pack of individual actuators. This assumed relationship} is necessary because the original paper does not discuss the wall-thickness of the equivalent actuator.

\subsection{The effect of wall-thickness}

Eq.~\eqref{eq:F_thick} can be manipulated algebraically to provide insight into the volumetric effect of the thickness on the actuator force output. The equation is first normalized by the gauge pressure $P'$ and the expression in \eqref{eq:D_from_D0} is substituted for $D_0$
\begin{equation}
    \begin{aligned}
      \frac{F_{\text{thick}}}{P'} = &\frac{\pi}{4} D^2 \left(2 \csc^2\!{\theta} - 3\right) \\
      &+ \pi \left( D t_k \left(2 - \csc^2\!{\theta} \right) - {t_k}^2 \right) .\\
    \end{aligned}
    \label{eq:F_thick_over_P}
\end{equation}
See the appendix for the derivation of Eq.~\eqref{eq:F_thick_over_P}. Note that this form of the equation needs to be used with caution. The form in Eq.~\eqref{eq:F_thick} is parameterized by $D_0$, which does not change during actuation. The form in Eq.~\eqref{eq:F_thick_over_P} is parameterized by the instantaneous external diameter $D$ which is itself a function of the braid angle $\theta$ and changes over the course of the actuation.

The expression in Eq.~\eqref{eq:F_thick_over_P} can be further normalized by the external cross-sectional area $A$ of the actuator defined by 
\begin{equation}
    A = \frac{\pi}{4} D^2
\end{equation}
\begin{equation}
\begin{aligned}
\frac{F_{\text{thick}}}{P'A} = &\left(2 \csc^2\!{\theta} - 3\right) + 4 \left(2 - \csc^2\!{\theta} \right) \frac{t_k}{D}  - 4{\left( \frac{t_k}{D} \right)}^2 .
\end{aligned}
\label{eq:F_thicK_over_PA}
\end{equation}

\begin{figure}
    \centering
    \includegraphics{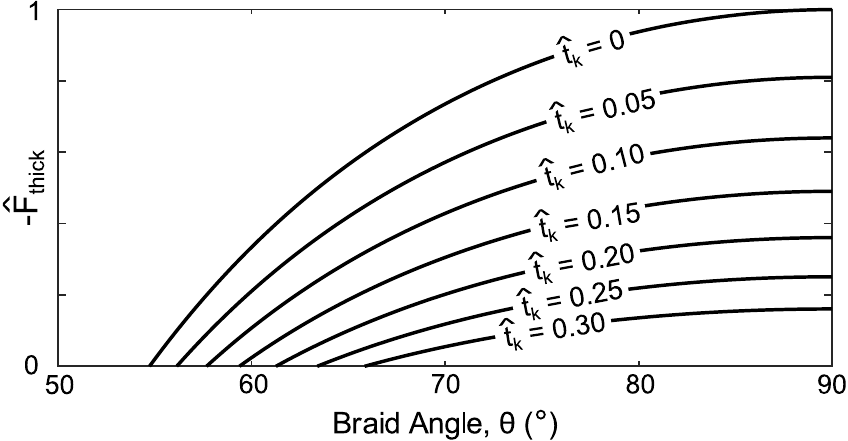}
    \caption{Shown is the relationship between the normalized extension force $\hat{F}_{\text{thick}}$, the braid angle $\theta$ and the relative elastomeric wall-thickness $\hat{t}_k$. The force shown is normalized by pressure and the current external cross-sectional area of the actuator. This idealized \hl{extension} force is highest at the theoretical braid angle of \unit[90]{$^\circ$} and decreases with the braid angle. An extending McKibben muscle, such as those considered in the original paper, is fabricated with some initial braid angle greater than \unit[54.7]{$^\circ$}. This angle decreases as the actuator extends. When there is no thickness, the maximum normalized force is one (equivalent to a piston with the same area). Increasing the relative thickness $\hat{t}_k$ decreases the maximum \hl{extension} force and increases the angle for which the force is predicted to reach zero.}
    \label{fig:F_hat}
\end{figure}

The normalized force in Eq.~\eqref{eq:F_thicK_over_PA} is designated $\hat{F}_{\text{thick}}$, a function of $\theta$ and the relative wall-thickness $\hat{t}_k$
\begin{equation}
\begin{aligned}
\hat{F}_{\text{thick}}\!\left(\theta, \hat{t}_k \right) = \left(-4\right){\hat{t}_k}^2 + \left(8 - 4\csc^2\!{\theta} \right)\hat{t}_k  +  \left(2 \csc^2\!{\theta} - 3\right) .
\end{aligned}
\label{eq:F_thicK_hat}
\end{equation}
This function is plotted in Fig.~\ref{fig:F_hat}. 

The predicted force output of an actuator at a given diameter is the product of the function in Eq.~\eqref{eq:F_thicK_hat} with the internal gauge pressure $P'$ and the external cross-sectional area $A$
\begin{equation}
F_{\text{thick}} =  \hat{F}_{\text{thick}}\!\left(\theta, \hat{t}_k \right) P' A.
\label{eq:F_thick_from_F_hat}
\end{equation}


\subsection{Considering the claims with wall-thickness}

Eq.~\eqref{eq:F_thick_from_F_hat} allows one to consider the claims of the original paper while considering the thickness of the elastomeric walls. The force $F_{\text{thick,pack}}$ from a pack of $n$ individual actuators is given by 
\begin{equation}
\begin{aligned}
 F_{\text{thick,pack}} &=  n F_{\text{thick,ind}} \\
&= n \hat{F}_{\text{thick}}\!\left(\theta, \hat{t}_{k\text{,ind}} \right) P' A_{\text{ind}} . 
 \end{aligned}
 \label{eq:F_thick_pack}
\end{equation}
The force $F_{\text{eq,thick}}$ from an actuator with an equivalent external cross-sectional area ($A_{\text{eq}} = n A_{\text{ind}}$) and the same relative wall-thickness ($\hat{t}_{k\text{,ind}} = \hat{t}_{k\text{,eq}}$) is given by
\begin{equation}
\begin{aligned}
 F_{\text{thick,eq}} &=  \hat{F}_{\text{thick}}\!\left(\theta, \hat{t}_{k\text{,eq}} \right) P' A_{\text{eq}} \\
  & =  \hat{F}_{\text{thick}}\!\left(\theta, \hat{t}_{k\text{,ind}} \right) P' \left(n A_{\text{ind}} \right) \\
  &= n \hat{F}_{\text{thick}}\!\left(\theta, \hat{t}_{k\text{,ind}} \right) P' A_{\text{ind}} .
 \end{aligned}
\label{eq:F_thick_eq}
\end{equation}

Eq.~\eqref{eq:F_thick_pack} is equivalent to Eq.~\eqref{eq:F_thick_eq}. Thus, even when considering the wall-thickness, the force from an equivalent actuator at the same pressure is predicted to be identical to the force from the actuator pack, regardless of the number of fascicles used
\begin{equation}
    F_{\text{thick,eq}} = F_{\text{thick,pack}}.
\end{equation}
This contradicts the claims of the original paper.

\subsection{Numerical verification}

As a final verification, the claims are considered through numerical experimentation. The force from an SPA pack made up from individual actuators with the parameters in Table~\ref{tab:table2} is calculated for a wall-thickness of \unit[1]{mm}. This is compared to the force from two equivalent actuators, one with the same \textit{relative} elastomeric wall-thickness and one with the same \textit{absolute} wall-thickness. This is repeated for various values of $n$ (i.e. various counts of individual actuators in the pack). All the calculated forces come from Eq.~\eqref{eq:F_thin} and MATLAB code for the verification is included in the appendix. The results are listed in Table~\ref{tab:num_results}.

\begin{table}[]
    \centering
    \resizebox{\columnwidth}{!}{\begin{tabular}{|c|c|c|c|} \hline
     $\;$     &  \textbf{SPA-Pack}  & \textbf{Equivalent Actuator}  &      \textbf{Equivalent Actuator} \\
     $n$      &  $nF_{\text{ind}}/P'$ &  Same \textit{relative} thickness  &  Same \textit{absolute} thickness  \\
      $\;$    &  ($t_{k,\text{ind}} = $ \unit[1]{mm})  & ($\hat{t}_{k,\text{eq}} = \hat{t}_{k,\text{ind}}$) & ($t_{k,\text{eq}} = t_{k,\text{ind}} = $ \unit[1]{mm}) \\ \hline
       1 & -0.165 & -0.165 & -0.165 \\
2 & -0.33 & -0.33 & -0.358 \\
4 & -0.661 & -0.661 & -0.755 \\
8 & -1.32 & -1.32 & -1.57 \\
16 & -2.64 & -2.64 & -3.22 \\
32 & -5.29 & -5.29 & -6.55 \\
64 & -10.6 & -10.6 & -13.3 \\ \hline
    \end{tabular}}
    \caption{Numerical Verification. The predicted force output per unit pressure (N/kPa). From Eq.~\eqref{eq:F_thick} and Table \ref{tab:table2}.}
    \label{tab:num_results}
\end{table}

As expected, the force from the pack is identical to that from the equivalent actuator with the same relative wall-thickness. This does not change with increasing numbers of individual actuators. The equivalent actuator with the same \textit{absolute} wall-thickness of \unit[1]{mm} shows the opposite trend of what is claimed by the original paper. That is, it creates a higher magnitude force than the pack and this disparity grows with an increasing $n$.

\section{Conclusion}

This paper demonstrates that the central claims of the popular original paper, ``Soft Pneumatic Actuator Fascicles for High Force and Reliability,'' \cite{fasciclesOriginal} are not true. The analysis of this paper refutes the original paper's claim that parallel combinations of extending McKibben muscles can produce more force-per-unit-volume than a single actuator of the same kind. It also refutes the original paper's claim that this benefit becomes more pronounced as the number of actuators in the pack increases. These claims have been refuted analytically, with the original paper's force model and with the law of conservation of energy. These claims have also been refuted by direct numerical calculation.




\section*{Appendix}

\subsection*{Derivation of Eq.~\eqref{eq:F_thick_over_P} from Eq.~\eqref{eq:F_thick}}

\begin{equation}
    \begin{aligned}
F_{\text{thick}} = &\frac{\pi P'}{4} {D_0}^2 \left( 3 \cos^2\!\left(\theta\right) - 1 \right) \\ 
    &+ \pi P' \left( D_0 t_k \left(2\sin{\theta} - \frac{1}{\sin{\theta}} \right) - {t_k}^2 \right) \\
    \\
     \frac{F_{\text{thick}}}{P'} = &\frac{\pi}{4} {D_0}^2 \left( 3 \cos^2\!\left(\theta\right) - 1 \right) \\ 
      &+ \pi \left( D_0 t_k \left(2\sin{\theta} - \frac{1}{\sin{\theta}} \right) - {t_k}^2 \right) \\
      \\
      \frac{F_{\text{thick}}}{P'} = &\frac{\pi}{4} {\left( \frac{D}{\sin{\theta}} \right)}^2 \left( 3 \cos^2\!\left(\theta\right) - 1 \right) \\ 
      &+ \pi \left( \frac{D}{\sin{\theta}} t_k \left(2\sin{\theta} - \frac{1}{\sin{\theta}} \right) - {t_k}^2 \right) \\
      \\
    \frac{F_{\text{thick}}}{P'} = &\frac{\pi}{4} D^2 \left( 3 \frac{\cos^2\!\left(\theta\right)}{\sin^2\!\left(\theta\right)} - \frac{1}{\sin^2\!\left(\theta\right)} \right) \\ 
      &+ \pi \left( D t_k \left(2 - \frac{1}{\sin^2{\theta}} \right) - {t_k}^2 \right) \\
      \\ 
      \frac{F_{\text{thick}}}{P'} = &\frac{\pi}{4} D^2 \left(2 \csc^2\!{\theta} - 3\right) \\
      &+ \pi \left( D t_k \left(2 - \csc^2\!{\theta} \right) - {t_k}^2 \right) .\\
    \end{aligned}
    \label{eq:F_thick_over_P_derivation}
\end{equation}

\vspace{1 cm}
 \newpage
 
\subsection*{MATLAB code for numerical verification}
\begin{verbatim}
close all
clear all

n_vec = [1,2,4,8,16,32,64]

L = .145; 
D_ind = 17e-3; s
theta = 80.3693712115*pi/180;
N = 16;
b = 0.866728223; 
t_k_ind = 1E-3; 
D_0_ind = b/(N*pi);

F_thick_over_P = @(D_0,t_k) ...
(pi/4)*D_0^2*(3*cos(theta)^2 - 1) ...
 + pi*(D_0*t_k*(2*sin(theta) - ...
 1/sin(theta)) - t_k^2)

F_over_P_ind = ...
F_thick_over_P(D_0_ind,t_k_ind);

for i = 1:length(n_vec)

  n = n_vec(i); 
  D_0_eq = sqrt(n)*D_0_ind; 

  t_k_eq1 = sqrt(n)*t_k_ind; 
  t_k_eq2 = t_k_ind;  


  F_over_P_pack(i) = F_over_P_ind*n;
  
  F_over_P_eq1(i) = ...
  F_thick_over_P(D_0_eq,t_k_eq1);
  
  F_over_P_eq2(i) = ...
  F_thick_over_P(D_0_eq,t_k_eq2);

  disp([num2str(n), ...
  ' & ',num2str(F_over_P_pack(i)*1000,3),...
  ' & ',num2str(F_over_P_eq1(i)*1000,3),...
  ' & ',num2str(F_over_P_eq2(i)*1000,3),...
  ' \\'])
end
\end{verbatim}
\end{document}